\documentclass{article}

\usepackage{microtype}
\usepackage{graphicx}
\usepackage{booktabs} %

\usepackage{caption}
\usepackage{subcaption}

\usepackage[hyphens]{url}

\usepackage{hyperref}

\usepackage{tikz}
\newcommand*\circled[1]{\tikz[baseline=(char.base)]{
            \node[shape=circle,draw,inner sep=1pt] (char) {#1};}}

\usepackage[accepted]{mlsys2021}

\mlsystitlerunning{FlexShard: Flexible Sharding for Industry-Scale
Sequence Recommendation Models}

\usepackage[compact]{titlesec}
\titlespacing{\section}{0pt}{1ex}{1ex}
\titlespacing{\subsection}{0pt}{1ex}{0ex}

\begin{document}

\twocolumn[
\mlsystitle{FlexShard: Flexible Sharding for Industry-Scale \\
Sequence Recommendation Models}

\mlsyssetsymbol{equal}{*}

\begin{mlsysauthorlist}

\mlsysauthor{Geet Sethi}{stanford,meta}
\mlsysauthor{Pallab Bhattacharya}{meta}
\mlsysauthor{Dhruv Choudhary}{meta}
\mlsysauthor{Carole-Jean Wu}{meta}
\mlsysauthor{Christos Kozyrakis}{stanford}

\end{mlsysauthorlist}

\mlsysaffiliation{stanford}{Stanford University, Stanford, California, USA}
\mlsysaffiliation{meta}{Meta, Menlo Park, California, USA}

\mlsyscorrespondingauthor{Geet Sethi}{geet@cs.stanford.edu}

\mlsyskeywords{Machine Learning, MLSys}

\vskip 0.3in

\begin{abstract}

Sequence-based deep learning recommendation models (DLRMs) are an emerging class of DLRMs showing great improvements over their prior sum-pooling based counterparts at capturing users' long term interests. These improvements come at immense system cost however, with sequence-based DLRMs requiring substantial amounts of data to be dynamically materialized and communicated by each accelerator during a single iteration. To address this rapidly growing bottleneck, we present \textit{FlexShard}, a new tiered sequence embedding table sharding algorithm which operates at a \textit{per-row granularity} by exploiting the insight that \textit{not every row is equal}. Through precise replication of embedding rows based on their underlying probability distribution, along with the introduction of a new sharding strategy adapted to the heterogeneous, skewed performance of real-world cluster network topologies, FlexShard is able to significantly reduce communication demand while using \textit{no additional memory} compared to the prior state-of-the-art. When evaluated on production-scale sequence DLRMs, FlexShard was able to reduce overall global all-to-all communication traffic by \textit{over 85\%}, resulting in end-to-end training communication latency improvements of \textit{nearly 6x} over the prior state-of-the-art approach.
\end{abstract}
]

\printAffiliationsAndNotice{}  %

\section{Introduction}
\label{intro}

Over the last decade the use of deep learning (DL) has become ubiquitous. Deep learning applications span a breadth of domains~\cite{imagenet,devlin-etal-2019-bert,hazelwood:2018:mlatfb,resnet,AlphaFold,naumov2020deep,transformers,Weyn_2020}, forming the backbone of many internet-scale services such as web search, social media, and video streaming~\cite{Cheng:dlrs2016, covington:2016:youtuberec, netflix,gupta:2020:archimp, tpu, Raimond:netflix, googe:2016:rankbrain, what_to_watch, zhou2019deep,wu:arxiv:2021}. One important class of deep learning applications which 
underlies
these services are recommendation systems. Modern recommendation systems are based on deep learning recommendation models to personalize contents for users in order to enhance the quality of experience. Deep learning recommendation models (DLRMs) constitute a significant and growing portion of AI compute cycles used in industry datacenters. For example, in 2019 and 2020, over 50\% of Facebook's AI training usage and over 80\% of its AI inference usage being used for DLRMs~\cite{gupta:2020:archimp,acun:2021:understandingtraining}.  

Unlike deep learning architectures used for tasks such as image recognition or language translation, which are comprised primarily of layers composed of general matrix multiplies (GEMMs) and exhibiting high compute-intensity per parameter, the majority of parameters making up DLRMs are contained in \textit{embedding layers}~\cite{naumov2020deep,Anil:recsys2022}. Embedding layers effectively function as massive parallel look-up tables exhibiting high memory intensity~\cite{gupta:2020:archimp,rec-nmp}. Recent works have shown that the memory storage requirement for DLRM embedding tables has grown by over 16 times between 2017-2021, with the latest models containing \textit{trillions} of parameters representing 100s of billions of embeddings requiring terabytes (TBs) of storage to hold the model weights. In addition, the memory bandwidth demands for DLRMs has increased by almost 30 times in the same time frame, requiring TBs of data to be read in a single training batch~\cite{mudigere2021highperformance, sethi2022recshard}.

In this work we propose a new approach to solve the \textit{sharding} problem for emerging \textit{sequence-based} DLRMs. In order to achieve the training throughout required for hyper-scale DLRMs with petabytes of training data~\cite{mudigere2021highperformance, zhao2022dataingestion}, GPUs with High Bandwidth Memory (HBM) have been deployed to accelerate DLRM training. However the latest GPUs contain HBMs on the order of 10s of gigabytes (GBs) --- orders-of-magnitude below the total memory footprint required to train these models, requiring systems to scale to distributed multi-GPU training to achieve the necessary throughput and introducing the sharding problem. The sharding problem is tasked with the partitioning and placement of embedding parameters across the system topology to maximize resource utilization and efficiency. What makes the sharding problem more difficult than applying the hand-tuned or automatically generated combinations of data, model, and/or pipeline parallelism present in many DL frameworks~\cite{tensorflow, pytorch} is the \textit{irregular and input data-dependent} access to embedding parameters.

\begin{figure}[t]
    \centering
    \begin{subfigure}[b]{0.45\textwidth}
  \includegraphics[width=\textwidth]{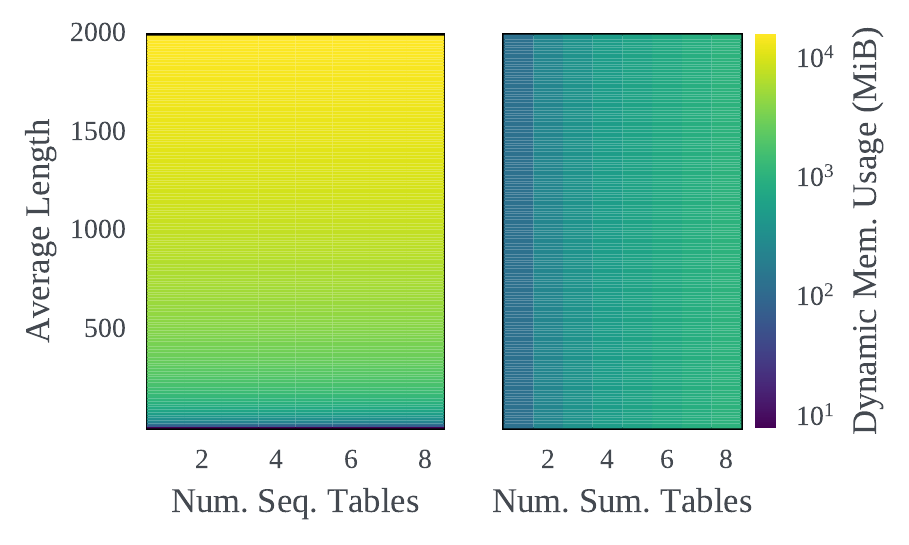}
\end{subfigure}
\begin{subfigure}[b]{0.45\textwidth}
  \includegraphics[width=\textwidth]{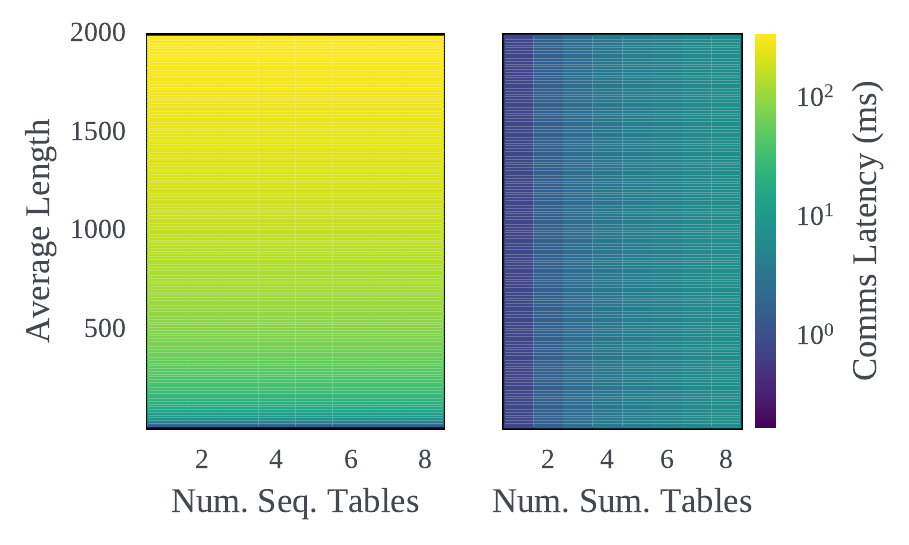}
\end{subfigure}
    \caption{Expected per-GPU memory and communication costs for a DLRM with up to 8 user-history embedding tables as total history length varies from 1 to 2000, where the features are either consumed as a sequence (left-column) or sum-pooled (right-column). Both variations are sharded row-wise by the SoTA TorchRec~\cite{torchrec} framework on a system with 4 nodes containing 32 GPUs, a local batch size of 4096, and rows of embedding dimension 256 and FP32 type. Sum-pooled embedding table memory and communication costs scale by the number of tables and \textit{do not vary} based on the feature length, while sequence embedding table costs \textit{scale linearly with feature length} and not by the number of tables.}
  \label{fig:sum_v_seq_scaling_mot}
       \vspace{-0.5cm}
\end{figure}

Furthermore, while the sharding problem has been increasingly explored in recent works due to its importance~\cite{adnan:2021:hotsplit, lui:2021:capacity, sethi2022recshard, autoshard}, they all, to our knowledge, assume that the embedding tables to be sharded are either \textit{one-hot}--meaning at most one embedding row per table will be accessed per training sample--or \textit{sum-pooled}--meaning \underline{all} embedding row accessed within a table by a training sample will be aggregated via summation before proceeding through the model. This underlying assumption can have substantial impact on the amount of \textit{dynamic} (i.e. runtime) memory 
usage and interconnect communication bandwidth needed by model training. Moreover, this assumption does not apply to state-of-the-art, emerging learned methods such as Transformers, where embeddings are consumed as a sequence to greatly improve recommendation accuracy~\cite{chen:seq_tranformers:seq, kang:self_attentive:seq, li:multi_interest:seq, pi:sequential_modeling, sheng:one_model:seq, sun:bert4rec:seq}. Figure \ref{fig:sum_v_seq_scaling_mot} illustrates the rapidly growing memory and communication costs of sequence-based embeddings, orders of magnitude greater than sum-pooling the same set of embeddings.

To address the rapidly growing training system bottleneck and to enable large-scale next generation sequence-based DLRMs, we propose \textit{FlexShard} — a new tiered sharding approach which partitions and places sequence DLRM embedding tables at a \textit{per embedding row granularity}. FlexShard is built upon the insight that \textit{not every row is equal} with respect to their system demands, and that careful, precise replication of certain rows can result in savings of \textit{both time and memory} compared to the existing  state-of-the-art industry sequence sharding strategy. Furthermore, FlexShard leverages the knowledge that real-world hardware network topologies are often heterogeneous, with non-equal bandwidth across all pairs of accelerators, to improve communication performance even further. Overall, we summarize the key contributions of this work as follows:
\begin{itemize}
    \item We develop an analytical framework to model the memory and communication scaling for existing sequence embedding sharding strategies, and show the potential for memory and communication savings through finer-grain sharding.
    \item We introduce a new sharding strategy—\textit{Flex}—which adapts to skewed real-world network interconnect (i.e. intra- vs. inter- node bandwidth) and communication collective (i.e. all-reduce vs. all-to-all) performance.
    \item We propose FlexShard -- a new multi-tiered sequence sharding algorithm which replicates and shards embedding tables at a fine, \textit{per-row granularity}. 
    \item Evaluation on production-scale sequence DLRMs, demonstrating that FlexShard can reduce global all-to-all communication traffic by \textit{over 85\%} compared to the baseline state-of-the-art approach~\cite{torchrec}, resulting in end-to-end training communication latency improvements of \textit{nearly 6x}. And FlexShard does so while also \textit{reducing peak memory utilization}.
\end{itemize}

\section{Background}
\label{background}

\begin{figure}[t]
    \centering
    \includegraphics[width=\linewidth,keepaspectratio]{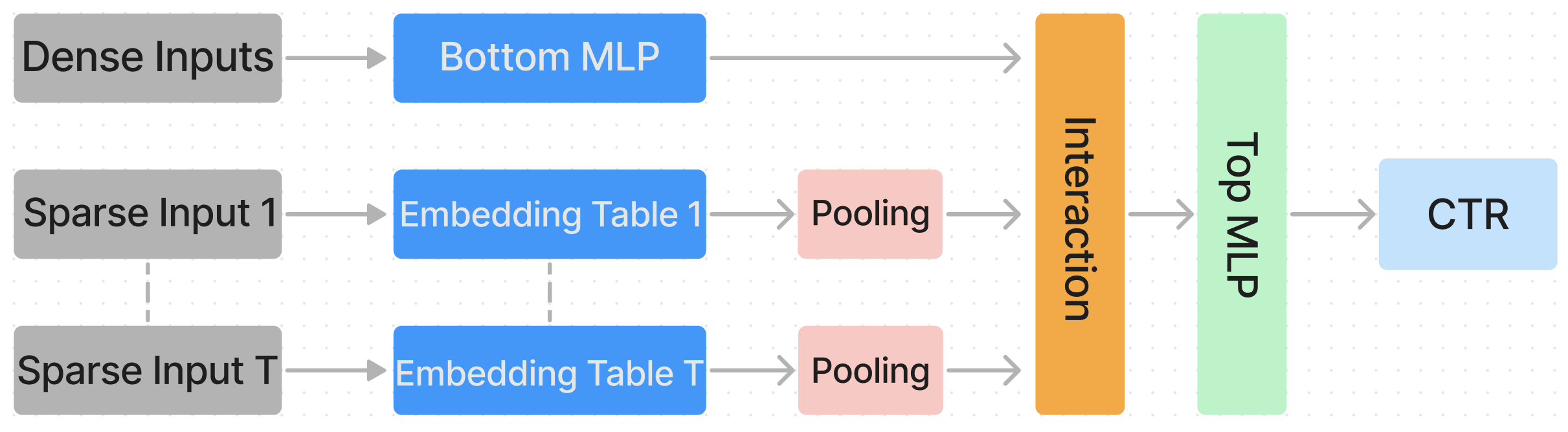}
    \caption{Canonical DLRM Architecture.}
  \label{fig:dlrm_overview}
  \vspace{-0.5cm}
\end{figure}

\textbf{DLRMs.} Figure \ref{fig:dlrm_overview} provides an overview of the generalized DLRM architecture which is composed of four primary components: embedding layers, pooling functions/layers, multi-layer perceptrons, and interaction functions/layers~\cite{naumov2019deep}. 

DLRMs solve the recommendation problem and provide personalized recommendations by predicting the probability that a user will interact with a particular piece of content, often referred to as the click-through-rate (CTR). To generate these predictions, DLRMs take as input two types of features representing both the user and content: \textit{dense} and \textit{sparse}. Dense features represent continuous input information and are consumed by the bottom-MLP layers, while sparse features represent discrete, categorical data and are passed as input IDs to the embedding layers. The function of the embedding layers is to transform these discrete IDs into learned, dense, continuous-valued \textit{embedding vectors} which encode the latent representation of each ID to be consumed by the rest of the model.   

For industry-scale recommendation systems, it is not uncommon for the cardinality of sparse features to reach the order of 100s of millions to billions, with each distinct ID representing, for example, individual videos on YouTube. Due to this scale, the embedding tables encoding these features can translate to 10s to 100s of gigabytes of storage requirements each~\cite{kang2021learning, mudigere2021highperformance, sethi2022recshard, zhao:2020:distributedgpu}. Furthermore, individual categorical features commonly encode multiple pieces of information simultaneously for each user, such as the history of IDs for content recently viewed, liked, shared, and so-forth by the user; each potentially being its own feature. As dense neural network (NN) layers such as MLPs require a fixed-sized tensor as input, the varying lengths of histories between users, each gathering a variable number of embedding vectors as input, must be aggregated and reduced in some fashion to a fixed-size output embedding, often referred to as the \textit{pooled embedding}. Once pooled, the various embedding outputs along with the outputs of the bottom-MLP layers undergo \textit{feature interaction} via a function such as the dot-product, before proceeding through the top-MLP layers and predicting the CTR.

\textbf{DLRM Sharding.} As DLRMs often contain multiple embedding tables, resulting in storage capacity demands greater than the available memory on any individual GPU, the training of DLRMs requires the use of \textit{model parallelism} and the solving of the \textit{sharding problem.} The goal of the sharding problem is to generate a \textit{balanced and efficient} partitioning and placement of embedding tables across the system topology such that each GPUs total execution time in each embedding related stage is \textit{as close to equal as possible.} This balance is of the utmost importance for hyper-scale DLRM training as these DLRMs are often trained \textit{synchronously}, where every instance of every model weight is identical across the entire system within the forward pass of a training batch, and whose gradient updates are globally synchronized at the end of each iteration before proceeding to the next~\cite{mudigere2021highperformance, sethi2022recshard}.

What makes the sharding problem more challenging than simply statically partitioning the weights of all embeddings is the \textit{balancing of compute and communication}. Each embedding table's compute requirement is unique and dependent on its input IDs—which are irregular and data-dependent—and its pooling function. Furthermore, the use of model parallelism requires the communication and distribution of embeddings across GPUs for potentially every batch to satisfy the local data dependencies of each GPU.

\textbf{Embedding Pooling.} A simple and common way to perform pooling is to do element-wise summation across all embeddings gathered to generate the output pooled embedding; known as \textit{sum-pooling}~\cite{deeprecsys, rec-nmp, kwon:scratch_pipe:sum, mudigere2021highperformance, sethi2022recshard, autoshard, zhou:deep_interest:sum, wilkening:2021:recssd}. The use of sum-pooling to perform embedding aggregation comes with beneficial properties with respect to the sharding problem. First, on modern hardware summation has very low compute intensity per byte, reducing the balancing of compute to simply the balancing of memory bandwidth demand (i.e. the number of IDs simultaneously accessed). Second, summation is a commutative and associative operation, allowing embeddings to be sharded arbitrarily and partially aggregated locally without impacting correctness. Third, the pooled output of each embedding table shard is always equal to the size of a single embedding row, and can be generated in-place without intermediate materialization all of the rows involved. Fourth, summation allows the use of high-performance communication collectives such as reduce-scatter, which can pool partially aggregated outputs \textit{while performing embedding distribution}.  

Summation however, disregards the \textit{sequential} nature of a user's history and behavior and places equal weight over all of a user's interactions over a time horizon. For example, for video streaming services such as Netflix and YouTube, it makes intuitive sense that the time ordered history of a user is of non-trivial importance to the content the user may want to engage with next. Therefore, similar to the history and progression of natural language processing (NLP) methods, evolving from bag-of-words methods to sequential, sentence based methods, it has recently been shown that by providing all of a user's embeddings to a \textit{learned} aggregation method, such as pair-wise self-attention and Transformers, can significantly improve recommendations~\cite{chen:seq_tranformers:seq, kang:self_attentive:seq, li:multi_interest:seq, pi:sequential_modeling, sheng:one_model:seq, sun:bert4rec:seq}.  

To our knowledge however, prior works which have developed novel solutions to the DLRM sharding problem have done so under the assumption that the embeddings gathered will all be sum-pooled before being consumed by the following model layers~\cite{adnan:2021:hotsplit, lui:2021:capacity, sethi2022recshard, autoshard}.
Unfortunately, these benefits do not directly apply to such sequence-based pooling methods where each embedding interacts with every other embedding in the sequence, requiring the full sequence to be present on the requesting GPU to perform the operation. Thus, each embedding must be materialized on the sending GPU to be communicated and sent to the requesting GPU, requiring a \textit{dynamic} amount of memory usage \textit{proportional} to the sequence length on the sending GPU. And furthermore, sequence embeddings cannot be reduced in-flight, requiring \textit{dynamic} memory equal to the sequence length on the requesting GPU, as well as the use of the slower all-to-all communication collective. 

Moreover, the lack of the benefits inherent to sum-pooling introduces a problem not discussed in prior work, and beyond the sharding axes they've explored for embedding tables: \textit{static} memory usage, 
lookup time, and communication time. This problem is that of \underline{\textit{dynamic}} memory 
and communication usage, 
which for current, trivially generated sequence lengths for large-scale internet companies (e.g. interactions with last 500 items served in the past 30 days), can result in dynamic memory usage \textit{much greater than} the corresponding static embedding table shard size on each GPU. Thus, dynamic memory utilization, in addition to the performance axes used in prior works, must also be considered in order to generate an efficient and performant sequence embedding sharding.

\textbf{Embedding Access Patterns}. Beyond the varying sizes and average compute demand (i.e. differing average pooling sizes) that occur across embedding tables, another important and distinct property which separates embedding layers not only from other NN layers but also from one another, is that of their \textit{sparse, irregular} access pattern~\cite{adnan:2021:hotsplit, kwon:scratch_pipe:sum, sethi2022recshard, wilkening:2021:recssd}. The amount of embedding rows accessed within a single data sample is many orders-of-magnitude less than the total number of rows in the table, and the rows accessed typically exhibit a power-law distribution, with a small percentage of the rows constituting an overwhelming majority of total accesses. While infrequently accessed, the long-tail cannot be ignored due to its importance in recommendation quality and user experience~\cite{zhao:2020:distributedgpu}. The presence of this \textit{access distribution skew} presents interesting opportunities for system-level optimizations.
\section{Black Box Sharding}
\label{bb_sharding}

Currently sequence embedding tables are treated as monolithic black box units by prevailing sharding algorithms~\cite{torchrec, mudigere2021highperformance}, which exploit information at the complete table and/or feature level, such as the feature's average length (i.e. pooling size). This level of detail does not provide granular information about the characteristics of the rows composing each table, requiring the strategies used by these algorithms to treat the memory access pattern \textit{within} each table as random.

\subsection{Uniform Strategies}

A class of strategies to shard tables using this level of information is to do so \textit{uniformly} across all ranks, through strategies such as column-wise (CW) and row-wise (RW), which disjointly partition the table model-parallel (MP); or data-parallel (DP), which replicates the table. CW shards tables by splitting embedding tables by their embedding dimension—i.e. columns—uniformly balancing the access load as every lookup is equally striped across all shards. Not as intuitive, a potentially surprising result is that sharding a table uniformly RW—that is, splitting the table into equal sized, \textit{contiguous} blocks of rows—also \textit{approximately} evenly distributed the lookup load, with highly accessed rows being spread across all shards. This nice result is due to raw content IDs—which generally are arbitrarily assigned 64-bit integers—being mapped to their corresponding embedding IDs via integer hash functions which uniformly distribute keys over the hash space—i.e. the embedding table size. Under DP, the embedding table is replicated across all GPUs, uniformly balancing load as each GPU performs all the lookups to the table from its local batch.

When sharded uniformly across \textit{all GPUs}, the memory access load with respect to the number of rows accessed in a global batch by \textit{each GPU} is equivalent in expectation across all uniform strategies. However, as we will explore in the following section, for other important properties such as: memory footprint, communication load, and input data distribution, the strategies can differ greatly.

\subsection{Modeling Black Box Performance}
\label{modeling_black_box}

In order to properly understand the scaling properties of existing black box uniform strategies and set a foundation upon which to develop more performant strategies, we begin by first building an analytical model of the expected work done by \textit{each GPU} under each black box strategy for a single embedding table in a single training iteration across 6 different metrics: 
\begin{enumerate}
    \itemsep-0.1em
  \item Static memory cost (bytes): the memory needed to store the shard of embedding table rows 
  \item Input Data Distribution Cost: the expected number of
IDs to be communicated before lookups can occur
  \item Rows accessed: the expected number of rows to be looked-up from the local embedding table shard 
  \item Dynamic memory cost (bytes): the expected runtime memory usage needed to store materialized lookup values and communication buffers; 
  occurs in both the forward and backward passes
  \item Dynamic communication cost (latency): the expected dynamic communication latency based on the max of the expected number of bytes communicated to and from the GPU and the interconnect performance;
  occurs in both the forward and backward passes
  \item Static communication cost (latency): the communication latency that occurs regardless of the batch and feature properties; 
  occurs in only the backward pass
\end{enumerate}

\begin{table}[t]
\normalsize
\centering
\resizebox{.9\columnwidth}{!}{%
\begin{tabular}{cl} 
\hline
\textbf{Parameter}                    & \textbf{Description}                      \\ 
\hline
\textit{E}                            & Number of Rows                            \\
\textit{D}                            & Embedding Dimension                            \\
\textit{dtype}     & Data-type of embedding scalars       \\
\textit{L}           & Average feature length                  \\
\textit{W}           & Number of GPUs in a node                  \\
\textit{N}      & Number of Nodes                         \\
\textit{U}      & Total number of GPUs ($=N*W$)                    \\
\textit{B}       & Local batch size per GPU   \\
\textit{a2a\textsubscript{global}} & Global All-to-All bandwidth \\
\textit{a2a\textsubscript{intra}}  & Intra-node All-to-All bandwidth                \\
\textit{ar\textsubscript{global}}  & Global All-Reduce bandwidth                \\
\textit{ar\textsubscript{cross}}       & Cross-node All-Reduce bandwidth     \\
\hline
\end{tabular}
} 
\caption{Parameters used in sharding scaling equations.}
\label{table:analytical_params}
\end{table}

\begin{table}[]
\centering
\resizebox{\linewidth}{!}{%
\begin{tabular}{|c|c|c|c|}
\hline
Sharding Type & Static Memory & Rows Accessed & Dynamic Memory   \\ \hline
& & &\\[-1em]
RW   & $\frac{E}{U}*D*dtype$ & $U*B*\frac{L}{U}*D$ & $2*B*L*D*dtype$\\ [.2em] \hline
& & &\\[-1em]
CW   & $E*\frac{D}{U}*dtype$ & $U*B*L*\frac{D}{U}$ & $2*B*L*D*dtype$\\ [.2em] \hline
& & &\\[-1em]
DP   & $6*E*D*dtype$ & $B*L*D$ & $B*L*D*dtype$\\ [.2em] \hline
& & &\\[-1em]
Flex   & $6*\frac{E}{W}*D*dtype$ & $W*B*\frac{L}{W}*D$ & $2*B*L*D*dtype$\\ [.2em] \hline
\end{tabular}}
\caption{Per-GPU expected memory costs for a sequence-based embedding table under different sharding strategies. Note, RW, CW, DP, and Flex stand for Row-Wise, Column-Wise, Data-Parallel, and our proposed design, respectively.}
\label{table:mem_scale}
\end{table}

\begin{table}[]
\centering
\resizebox{\linewidth}{!}{%
\begin{tabular}{|c|c|c|c|c|}
\hline
Sharding Type & Input Data Dist & Rows Accessed & Dynamic Comms & Static Comms   \\ \hline
& & & &\\[-1em]
RW   & $B*L$ & $U*B*\frac{L}{U}*D$ & $\frac{B*L*D*dtype}{a2a\_global}$ & N/A \\ [.2em] \hline
& & & &\\[-1em]
CW   & $U*B*L$ & $U*B*L*\frac{D}{U}$ & $\frac{B*L*D*dtype}{a2a\_global}$ & N/A \\ [.2em] \hline
& & & &\\[-1em]
DP   & N/A & $B*L*D$ & N/A & $\frac{E*D*dtype}{ar\_global}$ \\ [.2em] \hline
& & & &\\[-1em]
Flex   & $B*L$ & $W*B*\frac{L}{W}*D$ & $\frac{B*L*D*dtype}{a2a\_intra}$ & $\frac{\frac{E}{W}*D*dtype}{ar\_cross}$ \\ [.2em] \hline
\end{tabular}}
\caption{Per-GPU expected communication costs for a sequence-based embedding table under different sharding strategies.}
\label{table:comm_scale}
\end{table}

With these metrics defined, we can derive the scaling equations for each of these axes for every uniform strategy as shown in Tables \ref{table:mem_scale} and \ref{table:comm_scale}, with Table \ref{table:mem_scale} showing the memory scaling, Table \ref{table:comm_scale} showing the communication scaling, and Table \ref{table:analytical_params} summarizing the parameters used in the equations. 

Although the black box scaling equations are derived using coarse-grain information these strategies use—particularly the table size, $E$, for static costs, and the feature's average length, $L$, for dynamic costs—they provide compelling insights into the trade-offs between the strategies. 

One way to model the trade-offs between a pair of strategies is to form a ratio between their scaling equations, calculating the relative cost of one strategy normalized against another, as we vary a shared variable amongst the two. The uniform strategies can be cleanly separated into two groups based on how they treat parameters, either DP—with only \textit{all\_reduce} based communication occurring in the backward pass—or MP (i.e. CW and RW)—with \textit{all\_to\_all} based communication occurring in both the forward and backwards pass. Furthermore, we can reduce the modeling of trade-offs between DP and MP strategies to simply modeling the trade-offs between DP and RW, as in expectation RW and CW are identical across all metrics except for input data distribution, where CW requires strictly more communication.

\begin{figure}[t]
    \centering
    \includegraphics[width=.9\linewidth,keepaspectratio]{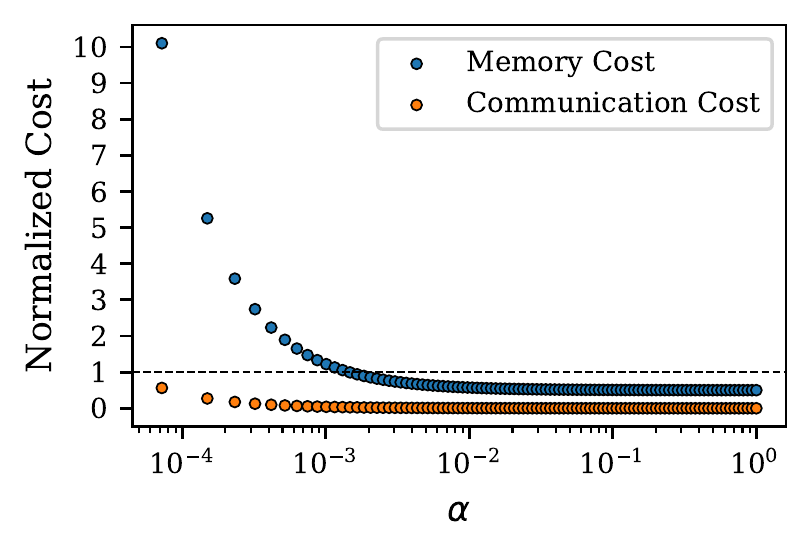}
    \vspace{-0.4cm}
    \caption{Normalized memory and communication costs of DP normalized to RW for a fixed embedding table size of $E$ and varying $L = \alpha E$. Target training configuration of 32 GPUs, batch size of 4096, embedding dimensions of 256, and global bandwidths $all\_reduce_{bw} = 3 \cdot all\_to\_all_{bw}$.}
  \label{fig:dp_v_rw_scaling}
  \vspace{-0.4cm}
\end{figure}

In Figure \ref{fig:dp_v_rw_scaling} we show the total normalized memory and communication costs of DP normalized to RW as we fix $E$ and vary $L$ as: 
\begin{equation} \label{eq:l_e_ratio}
L=\alpha \cdot E
\end{equation}
Furthermore, we plot the horizontal line $y=1$, which divides each curve into the regions where DP is either more efficient than RW (i.e. $y < 1$), or less efficient (i.e. $y > 1$). As communication costs are dependent on the target hardware topology (for their bandwidths), Figure \ref{fig:dp_v_rw_scaling} is modeled using the same topology as in our experiments, the details of which are in Section \ref{methodology}.

Due to the coarse-grain, table level decision making of existing black box algorithms, we highlight that \textit{in practice these approaches will almost never employ DP for sequence embedding tables}. This is because the high cardinality of sequence features leads to embedding table sizes on the order of 10s of millions of rows, and even average sequence feature lengths on the order of 1000s would result in $\alpha \sim O(10^{-4})$; for which, as shown in Figure \ref{fig:dp_v_rw_scaling}, DP memory requirements will be much greater than RW. Therefore existing approaches are unable to realize any of the communication benefits of DP.

However, Figure \ref{fig:dp_v_rw_scaling} also shows us that for sufficiently large ratios of average feature length to table size (i.e. $\alpha$), DP saves \textit{both time and memory} compared to RW. While, as previously mentioned, this is unlikely to occur at the whole table granularity, it may be possible to construct such a ratio using \textit{a subset of rows}. This insight inspires FlexShard, which by sharding at a per-row rather than table granularity, can attempt to leverage the best of both DP and RW.

\section{FlexShard}
\label{prob_sharding}

Building upon the insights so far, we now formulate and develop \textit{FlexShard}. FlexShard is a sharding algorithm which optimally shards sequence embedding tables at a \textit{per-row granularity} based on the underlying feature distributions. 

\subsection{Not Every Row is Equal}
\label{not_every_row}

\begin{figure}[t]
\centering
    \begin{subfigure}[b]{0.23\textwidth}
  \includegraphics[width=\textwidth]{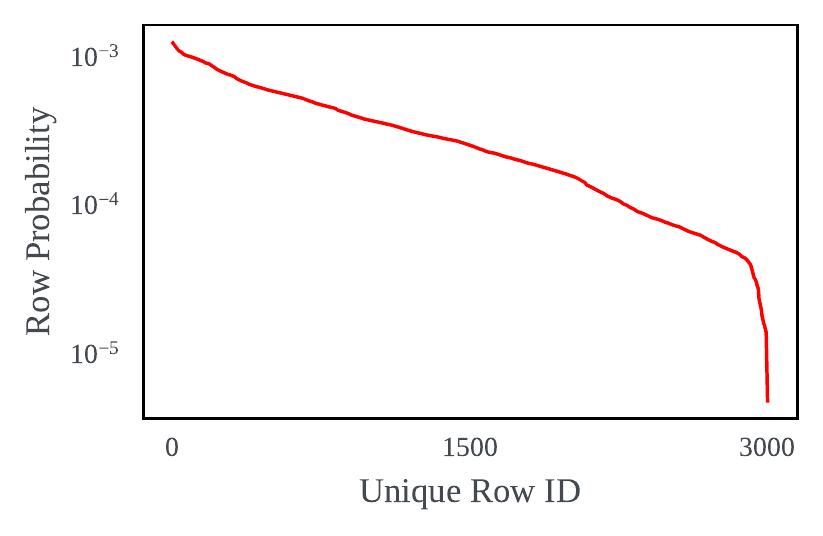}  
\end{subfigure}
    \begin{subfigure}[b]{0.23\textwidth}
  \includegraphics[width=\textwidth]{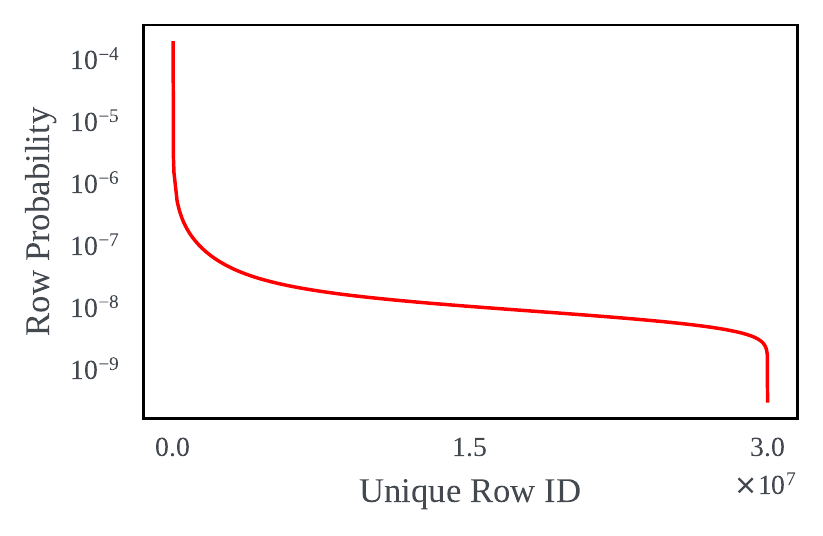}  
\end{subfigure}
    \begin{subfigure}[b]{0.23\textwidth}
  \includegraphics[width=\textwidth]{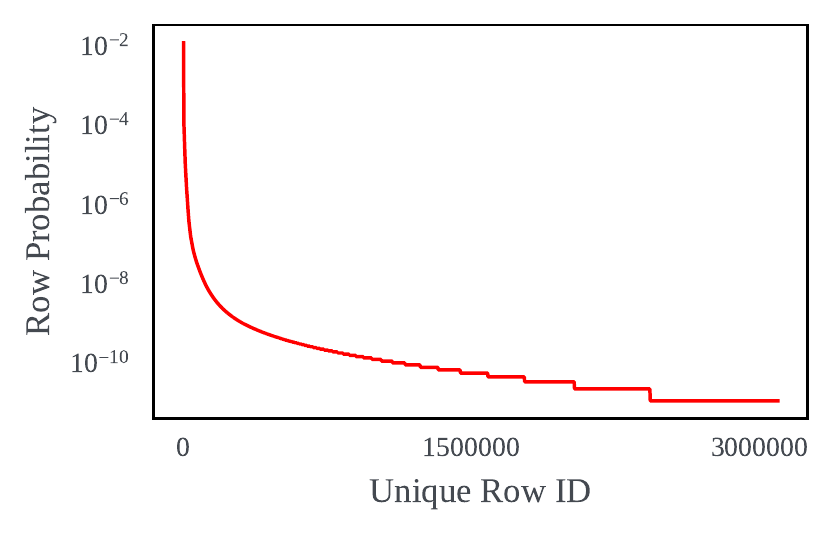}  
\end{subfigure}
    \begin{subfigure}[b]{0.23\textwidth}
  \includegraphics[width=\textwidth]{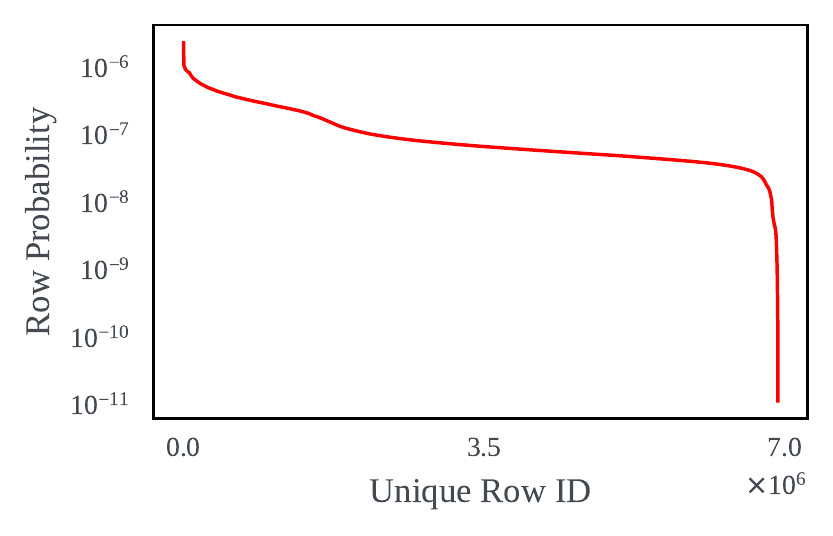}  
\end{subfigure}
    \caption{Ordered, normalized value frequency distributions for several sparse features used in production DLRMs. Despite the large variations of distribution shape, all distributions shown have the same average feature length. The distributions are generated from billions of randomly-selected training data samples.}
  \label{fig:value_freq_dists}
  \vspace{-0.5cm}
\end{figure}

Figure \ref{fig:value_freq_dists} shows the normalized sorted value frequency counts—\textit{that is, the cumulative number of appearances of each unique feature value}—of various sparse features over billions of production training data samples. While not identical, each follows the same general shape corresponding to an underlying power-law distribution,
where a small percentage of IDs represent the majority of accesses. 

Furthermore, although each distribution has a unique shape, \textit{they all have the same L}. Therefore each \textit{will be treated the same and sharded identically by black box strategies}. 

To address this clear limitation as well as provide an avenue to leverage our insight from the previous section, we formulate a definition for computing a feature's $L$, which, unlike prior and related works to our knowledge, is \textit{computed directly from its underlying value distribution}. 
Each point in the distributions in Figure \ref{fig:value_freq_dists} represents the probability that a particular row (embedding index) will be present in a random data sample of the feature. This equivalently corresponds to the row's binomial probability $p$, with its expected value of appearing in a random sample being $E[X_{i}]=p$. Under this formulation we then define the value distribution to be the set of all $E$—the number of rows—binomial random variables. 

With this definition we can then compute the expected number of rows to be present in a random data sample of this feature as 
\begin{equation} \label{eq:l_sum}
E[X]=\sum_{i=1}^{E}p_{i}=L
\end{equation}
due to the \textit{linearity of expectation}. Note that we \underline{do not} make any assumptions regarding the independence, or lack of, between rows. Or in other words, assumptions regarding \underline{any correlation} between particular rows being present. This is because the linearity of expectation \textit{does not require independence} between the random variables involved. 

Computing $L$ this way provides us with two important insights: first, we can calculate the $L$ of an arbitrary subset of rows based on their individual per-row probabilities; and second, every row \underline{does not} necessarily contribute the same weight, that is, \underline{not every row is equal}.

\subsection{Separating the Costs}

With this insight, let us now revisit the scaling equations and normalized cost models of Section \ref{modeling_black_box}. Substituting $L$ with the per-row based summation of eq. \ref{eq:l_sum} allows us to \textit{separate the total costs of a table into the summation of the individual per-row costs}, with the $L$ of each row being its per-row probability. Doing so, we can also substitute $E=1$ into eq. \ref{eq:l_e_ratio}, resulting in Figure \ref{fig:dp_v_rw_scaling} modeling the normalized costs of replicating \textit{a row} DP vs. RW sharding it, based on the row's probability. 

Furthermore, solving for the point on each curve intercepted by the line $y=1$ provides us with two alphas, or \textit{breakpoints}, which identify the row probabilities at which DP is either memory or communication \textit{neutral} with RW. And as a result, also identify the regions of the probability space for which DP'ing a row is more efficient than RW for either memory cost, communication cost, \textit{or both}.

\subsection{Sharding Row by Row}

\begin{figure}[t]
    \centering
    \includegraphics[width=.9\linewidth,keepaspectratio]{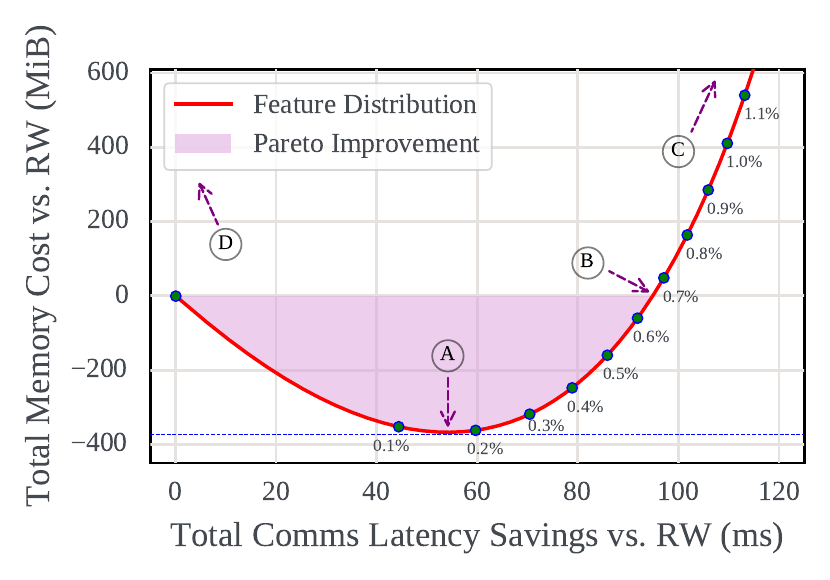}
    \caption{Expected memory and communication costs as increasingly more rows, sorted in descending frequency order, of a production sequence embedding table are replicated DP instead of RW sharded.}
  \label{fig:dp_v_rw_pareto}
  \vspace{-0.5cm}
\end{figure}

How can we use this to shard? As the breakpoints separate a table's value distribution into regions, if we traverse a table's value distribution in sorted frequency order, then across all breakpoints, the next row encountered will always be the most marginally efficient row (most positive, least negative) of all the rows remaining for that breakpoint. 

We show this in Figure \ref{fig:dp_v_rw_pareto}, where we plot the cumulative marginal memory (y-axis) and communication (x-axis) costs of replicating a row DP vs. RW sharding it, as we traverse an increasing number of rows of a sequence embedding table's value distribution in descending sorted frequency order (with the first 1.1\% of rows shown—marked in 0.1\% increments). This figure visualizes the simultaneous, joint marginal memory and communications costs of replicating specific subsets of rows DP while RW sharding the rest. Along the curve we identify four important points of interest based on the breakpoints and the underlying distribution.

The first point, \circled{A}, occurs at the minima of the curve, and identifies the sharding which generates the \textit{maximum memory savings}, and its corresponding communication savings. It occurs when the probability of the next row in the descending sorted value distribution is less than the memory-neutral breakpoint probability. From then onward, all remaining rows will incur a marginal memory cost versus RW. 

The second point, \circled{B}, occurs at the x-intercept and denotes when the cumulative marginal memory cost reaches zero. This second point identifies the \textit{memory-neutral} sharding, which generates \textit{the maximum amount of communication savings at no additional memory cost}. This sharding design is important to identify as it represents the maximal \textit{`drop-in'} throughput improvement for models and training topologies which may already exist and run using a baseline strategy such as RW. 
Furthermore, as the target training topology is fixed during sharding and we do not wish to under-utilize HBM, this is the target sharding design for the rest of the paper and evaluation, and is referred to as \textit{FlexShard 2-tier}. 

The third point, \circled{C}, occurs when the communication neutral breakpoint is reached. This marks the sharding which generates the \textit{maximum communication savings}, 
and after which all rows will incur a marginal communication cost vs. RW. Lastly, the final point, \circled{D}, provides the total marginal memory and communication costs of replicating the entire table DP vs. RW sharding it. 

While these points clearly identify potential sharding designs of interest based on the breakpoints and the underlying distribution, we highlight that \textit{every point along the curve} represents a potential sharding and that the curve itself is the \textit{Pareto frontier} for the memory and communication trade-off for per-row DP vs. RW sharding. And therefore, for any specified amount of total marginal memory cost, FlexShard can output the optimal hybrid DP/RW sharding, if it exists, with the minimal communication latency. 

\subsection{Adapting to Real-World Hardware Topologies}

The 2-tier FlexShard design developed thus far is built upon the same uniform strategies used by black box algorithms: DP and RW. An intrinsic property of these strategies are that they are \textit{globally uniform}, meaning that the amount of work done \textit{by and between} all GPUs is effectively identical. While easier to reason about, global uniformity is oblivious to the fact that real-world hardware topologies are often \textit{heterogeneous and tiered}. Four such examples of industry and cloud-scale training systems are Meta's ZionEX nodes~\cite{zionex, mudigere2021highperformance}, Azure's NDv4 VMs~\cite{azure:ndv4}, AWS's EC2 P4d instances~\cite{aws:p4d}, and NVIDIA's DGX nodes~\cite{nvidia:dgx}, each of which containing much higher \textit{intra}-node bandwidth through the use of NVLink/NVSwitch interconnects as compared to their \textit{inter}-node lower bandwidth RDMA based interconnects such as RoCE and InfiniBand.      

Reflecting on our learnings so far, we developed a new sharding strategy—\textit{Flex}—to leverage FlexShard's ability to shard at a per-row granularity based on the underlying 
distribution while concurrently adapting to the skewed communication bandwidth present in heterogeneous network topologies. 

To balance the exponentially increasing memory cost of DP'ing a row as the per-row probability further decreases with respect to the memory-neutral breakpoint, while leveraging higher intra-node bandwidth vs. inter-node bandwidth, the Flex strategy shards a set of rows RW \textit{within a node}, and then \textit{replicates each shard DP across nodes}. Doing so allows the Flex strategy to exploit the communication efficiency of DP across nodes, where bandwidth utilization is much more expensive, while simultaneously leveraging RW to reduce per-GPU memory usage, which is limited in capacity, by exploiting the much higher intra-node bandwidth available (Tables \ref{table:mem_scale} and \ref{table:comm_scale}).

\subsection{Sharding Across Three-Tiers}

To utilize the Flex strategy, we update the previously described 2-tier algorithm to \textit{FlexShard 3-tier} as follows. First, proceeding in descending sorted frequency order, all rows until the memory-neutral breakpoint (\circled{A} in Figure \ref{fig:dp_v_rw_pareto}) are replicated DP across all GPUs, generating memory and communication savings. Next, rows are sharded Flex until either all memory savings generated are consumed ((\circled{B} in Figure \ref{fig:dp_v_rw_pareto}), or until the communication-neutral breakpoint for Flex vs. RW is reached. Finally, all remaining rows are sharded RW. This results in the memory-neutral \textit{FlexShard 3-tier}.   

\subsection{Sharding Together, All at Once}

Due to the balanced and per-row design of FlexShard, it is trivial to extend FlexShard to multi-table sequence sharding. To do so, the probability distributions of all embedding tables are simply merged and sorted into one large distribution. FlexShard then just shards this distribution, optimally trading off the underlying distributions simultaneously at a precise row-wise granularity, without any additional input.

\section{Experiment Methodology}
\label{methodology}

\textbf{Baseline.} As we are, to our knowledge, the first work exploring the performance of sequence embedding sharding algorithms, we use as a baseline TorchRec~\cite{torchrec}, Meta's state-of-the-art open-source production sharding framework integrated with PyTorch~\cite{pytorch} which has support for sequence embeddings and implements the row-wise sharding strategy. 

\textbf{FlexShard.} To properly evaluate FlexShard and the proposed Flex sharding strategy, we implemented and compared two versions of FlexShard with the memory-neutral goal against the baseline; the first, 2-tier FlexShard, utilizing only two potential sharding strategies, DP and RW; the second, 3-tier FlexShard, utilizing all three strategies discussed in this work: DP, RW, and Flex.   

\textbf{Dataset.} There are, to our knowledge, no public industry-scale datasets designed for sequence DLRMs. Moreover, existing academic datasets released for sum-pooled DLRMs primarily contain 1-hot data (meaning \underline{at most 1 row} is accessed per embedding table per sample) and are designed for models which can wholly fit in the latest GPUs, making the sharding problem unnecessary. Therefore we evaluate this work by training four different industry-scale sequence DLRMs, each with a unique underlying feature distribution, on production ML training datasets.

\textbf{Training System Specification.} All experiments were performed on a cluster of 4 training nodes, each containing: 4-socket CPUs with 1.5TB of DRAM and 8x NVIDIA A100 40GB GPUs connected intra-node via high-bandwidth NVLink 3.0 in a hybrid cube mesh topology. For inter-node communication, all GPUs have a dedicated 200 Gbps RoCE NIC. NCCL was used as the communication 
library.

\textbf{Training Configuration and Implementation.} All models used were written in PyTorch, with their specifications detailed in Table~\ref{table:dlrm_specs}. The models were then wrapped, sharded, and trained using the TorchRec library, with both FlexShard versions implemented in CUDA/Python and integrated with PyTorch/TorchRec. The local batch size used for training was 4,096; resulting in a global batch size of 131,072.

\textbf{Performance Profiling.} As our goal is to assess the accuracy of FlexShard's analytical models as well as the communication latency improvement due to its multi-tier sharding design, rather than just end-to-end training throughput improvement, we perform our evaluation by tracing the execution of each GPU. To accurately do so, our tracing is executed synchronously across all 32 GPUs after 50 warmup iterations and is performed using the production libkineto profiling library integrated with the PyTorch Profiler. Profiling in this manner not only allows us to analyze per-kernel timing information, but also record and view additional information such as the payload size of each communication collective. To accurately measure memory utilization, we performed an additional experimental run using a version of PyTorch compiled with the CUDA Caching Allocator disabled and continuously monitoring each GPUs HBM utilization throughout training via metrics gathered using the NVIDIA Data Center GPU Manager~\cite{nvidia:dcgm}.

\begin{table}[]
\centering
\resizebox{\linewidth}{!}{%
\begin{tabular}{|c|c|c|c|c|c|}
\hline
Model & Sequence & Sequence & Embedding & Static & Local    \\
& Avg. Length & Table Size & Dimension & Size & Activation Size   \\ \hline
RM-1   & $\sim 1000$   & 30M  &      256      & 30 GB & 4.19 GB  \\ \hline
RM-2   & $\sim 1000$   & 30M  &      256      & 30 GB & 4.19 GB \\ \hline
RM-3   & $\sim 500$   & 10M  &      256      & 10 GB & 2.10 GB  \\ \hline
RM-4   & $\sim 500$   & 10M  &      256      & 10 GB & 2.10 GB  \\ \hline
\end{tabular}}
\caption{DLRM Sequence Embedding Specifications. 
}
\label{table:dlrm_specs}
\vspace{-0.5cm}
\end{table}
\section{Evaluation}
\label{eval}

We now present the evaluation results of FlexShard and its various contributions. First, we provide and discuss the output sharding by each of the FlexShard designs and its associated analytically expected communication improvements under the memory-neutral optimization goal. Next, we present and analyze the observed communication sizes and latencies for each communication collective across all three sharding designs; how they compare to the analytically expected values; and the resulting end-to-end throughput improvement. Lastly, we present and discuss the observed HBM utilization across all GPUs for each of the sharding designs and how they compare to one another.

\subsection{Generated Shardings and Expected Improvements}

\begin{table*}[]
\centering
\resizebox{.9\textwidth}{!}{
\begin{tabular}{c|c|c|c|c|c|c|c|c}
  Model & Sharding & DP Rows & Predicted DP  & Flex Rows & Predicted Flex & RW Rows & Predicted RW & Predicted Additional \\ 
  & & & Covered Access \% & & Covered Access \% & & Covered Access \% & Memory Cost \\ \hline
RM-1 & 2-tier & 346,144 & 52.5\% & N/A & N/A & 29,653,856 & 47.5\% & $\sim$ 0 \\  
RM-1 & 3-tier & 100,064 & 36.8\% & 1,184,544 & 30.6\% & 28,715,392 & 32.6\% & $\sim$ 0 \\ \hline
RM-2 & 2-tier & 507,072 & 77.0\% & N/A & N/A & 29,492,928 & 23.0\% & $\sim$ 0 \\ 
RM-2 & 3-tier & 128,736 & 63.9\% & 2,397,216 & 21.7\% & 27,474,048 & 14.4\% & $\sim$ 0 \\ \hline
RM-3 & 2-tier & 115,296 & 34.9\% & N/A & N/A & 9,884,704 & 65.1\% & $\sim$ 0 \\ 
RM-3 & 3-tier & 36,704 & 22.7\% & 318,560 & 25.9\% & 9,644,736 & 51.4\% & $\sim$ 0 \\ \hline
RM-4 & 2-tier & 184,864 & 57.0\% & N/A & N/A & 9,815,136 & 43.0\% & $\sim$ 0 \\ 
RM-4 & 3-tier & 55,040 & 39.0\% & 592,992 & 36.0\% & 9,351,968 & 25.0\% & $\sim$ 0 \\ \hline
\end{tabular}}
\caption{Number of rows placed in each sharding tier and their predicted access coverage by FlexShard 2-tier and 3-tier for each RM.}
\label{table:predicted_results}
\end{table*}

Table \ref{table:predicted_results} shows the number of rows placed in each strategy by each FlexShard version, along with the corresponding analytical percentage of total embedding communication volume covered. There are a few important observations present. First, while every DLRMs distribution has a region which can be used to generate memory-neutral communication savings, the size of this region is very small compared to the total table size, representing less than 1\% of the table. This is in part due to the high memory cost of DP'ing parameters in existing DL frameworks (the estimated $6x$ multiplier, taken from TorchRec), much greater than the `basic' $1x$ cost (on each GPU) required to simply replicate parameters. This is also further exemplified by the amount of rows able to be Flex sharded under 3-tier FlexShard  (which only maintains one replica per node as opposed to per GPU). Thus, the communications savings and throughput improvements of the memory-neutral goal will continue to increase, even while holding the model and training hardware constant, as the framework software improves.

Second, although the total amount of rows moved to either DP or Flex is relatively small compared to the raw table size, due to the underlying embedding input data being power-law distributed, we observe a predicted asymmetric coverage of total embedding communication volume by FlexShard as compared to the baseline RW. This allows 3-tier FlexShard for RM-2 to cover 85.6\% of the total all-to-all embedding communication volume with only .4\% of rows replicated DP and 8\% Flex sharded.

The insight of Section \ref{not_every_row}, that not every row is equal, is also demonstrated by comparing the generated sharding for RM-1 vs. RM-2, and RM-3 vs. RM-4. Although RMs-\{1,2\} and RMs-\{3,4\} each have (approximately) the same average length, and are thus treated identically by the baseline SoTA, FlexShard adapts each sharding to the underlying distributions, resulting in memory-neutral designs which differ both in the magnitude of rows placed in each strategy and the amount of accesses covered.

\subsection{Realized Communication and Throughput Improvements}

\begin{table*}[]
\centering
\resizebox{.9\textwidth}{!}{
\begin{tabular}{c|c|c|c|c|c|c|c}
  Model & Sharding & Global All-to-All & Global All-to-All & Global All-to-All  & Intra All-to-All & Global All-Reduce & Cross All-Reduce \\ 
  & & Size (each direction) & Reduction \% & Total Time & Total Time (Overlappable) & Total Time & Total Time (Overlappable)  \\ \hline
  
  RM-1 & RW & 972,424,960 & N/A & 307 ms & N/A & N/A & N/A \\ 
  RM-1 & 2-tier & 450,510,848 & 53.7\% & 149 ms & N/A & 3 ms & N/A \\ 
  RM-1 & 3-tier & 297,697,024 & 69.4\% & 99 ms & 30 ms & 3 ms & 14 ms \\ \hline
  RM-2 & RW & 958,656,768 & N/A & 308 ms & N/A & N/A & N/A \\ 
  RM-2 & 2-tier & 220,126,208 & 77.0\% & 80 ms & N/A & 6 ms & N/A \\ 
  RM-2 & 3-tier & 138,198,784 & 85.6\% & 49 ms & 21 ms & 3 ms & 27 ms \\ \hline
  RM-3 & RW & 466,826,496 & N/A & 156 ms & N/A & N/A & N/A \\ 
  RM-3 & 2-tier & 311,369,728 & 33.3\% & 101 ms & N/A & 4 ms & N/A \\ 
  RM-3 & 3-tier & 246,077,440 & 47.3\% & 85 ms & 13 ms & 2 ms & 5 ms \\ \hline
  RM-4 & RW & 473,709,312 & N/A & 155 ms & N/A & N/A & N/A \\ 
  RM-4 & 2-tier & 201,887,232 & 57.4\% & 72 ms & N/A & 5 ms & N/A \\ 
  RM-4 & 3-tier & 118,944,768 & 74.8\% & 45 ms & 18 ms & 3 ms & 9 ms \\ \hline
\end{tabular}}
\caption{Observed 
communication performance results of FlexShard 2-tier (DP/RW) and FlexShard 3-tier (DP/RW/Flex) versus the state-of-the-art row-wise baseline on RM1, RM2, RM3, and RM4.
}
\label{table:experimental_results}
\end{table*}

\begin{figure}[t]
    \centering
    \includegraphics[width=.9\linewidth,keepaspectratio]{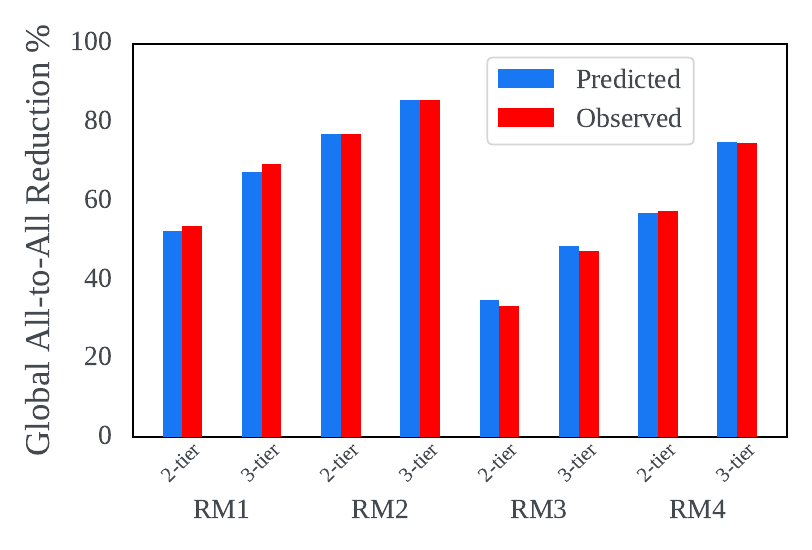}
    \vspace{-0.3cm}
    \caption{FlexShard Global All-to-All Reduction Percentage.}
  \label{fig:global_a2a_reduction}
  \vspace{-0.3cm}
\end{figure}

\begin{figure}[t]
    \centering
    \includegraphics[width=.9\linewidth,keepaspectratio]{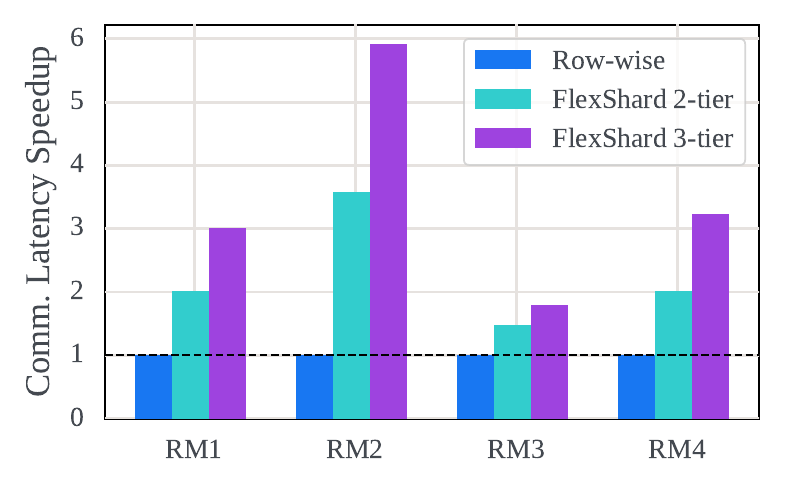}
    \vspace{-0.3cm}
    \caption{End-to-end training communication latency improvement of FlexShard 2-tier and 3-tier over the state-of-the-art row-wise baseline on RM1, RM2, RM3, and RM4.}
  \label{fig:speedup}
  \vspace{-0.5cm}
\end{figure}

Table \ref{table:experimental_results} summarizes the  results during model training, confirming the predicted improvements with large reductions in global all-to-all size as we move from the baseline RW-only sharding to the 2- and 3-tier FlexShard designs. In RM-1, for example, we see a global all-to-all reduction of 53.7\% and 69.4\% for FlexShard 2- and 3-tier, respectively, closely matching the analytically predicted reductions of 52.5\% and 67.4\%. These reductions come at the cost of additional smaller, faster, and potentially overlappable collectives. Together, this results in a 61\% and 81\% end-to-end throughput improvement respectively, for FlexShard 2-tier and 3-tier, over the baseline RW sharding for RM-1. 

In Figure \ref{fig:global_a2a_reduction} we plot the predicted vs. observed global all-to-all traffic reduction for both FlexShard designs across all RMs. What we can quickly observe is that the analytical models used by FlexShard are quite accurate at predicting the amount of communication costs based on the target topology and underlying feature distributions.

\subsection{Memory Utilization}

To accurately evaluate the empirical memory usage of FlexShard against the RW baseline we ran our experimental setup for RM-1 with the CUDA Caching Allocated disabled. This forces PyTorch to continuously allocate/deallocate GPU memory, allowing us to more accurately measure the true memory usage of each sharding strategy on the model. From this we observed that instead of being memory-neutral as intended, both FlexShard strategies \textit{actually reduced peak memory usage} by a non-trivial amount compared to the baseline RW sharding. Specifically, under the 2-tier design FlexShard uses approximately 2.5GiB less per GPU, and approximately 1.5GiB less per GPU when using the 3-tier design. The reason for this, albeit good, discrepancy is due to the DP memory overhead multiplier used in the analytical model. The $6x$ multiplier used, which is taken from TorchRec, actually overestimates the memory needed to DP a set of parameters. Therefore, the realized communication latency and throughput improvements of FlexShard presented occur while \textit{also lowering memory utilization.}

\section{Related Work}
\label{related_work}

The systems implications of DLRM training
is an area of increasing attention due to its real-world impact and unique challenges compared to more traditional DL architectures.

To tackle their size, works have proposed optimizations such as: per-row scaling of embedding dimension, based on the frequency of access~\cite{ginart2019mixed};  embedding table tensor decomposition and approximation~\cite{yin:2021:ttrec}; and even the replacement of whole embedding layers with their own deep neural networks~\cite{kang2021learning}. These optimizations introduce trade-offs, as they attempt to balance decreased memory capacity demand with increased compute requirements and potential accuracy impacts.

A separate category of optimizations target the access latency of embedding tables
These works have broadly focused on one of two approaches, either dynamic caching or the static placement (i.e. sharding) of embedding tables. ScratchPipe~\cite{kwon:scratch_pipe:sum} and RecShard~\cite{sethi2022recshard} tackle the problem of embedding access latency in hybrid CPU-GPU training systems
ScratchPipe addresses the problem through the use of a run-ahead GPU-side cache to attempt to have all embedding accesses hit in local GPU HBM, while RecShard uses a mixed-integer linear program and the per embedding table distributions to statically place the most frequently accessed rows in GPU HBM. AutoShard~\cite{autoshard} focuses on the sharding of embedding tables in a multi-GPU only training system, and uses deep reinforcement-learning and a neural network based cost model to perform its placement decisions. 

While similar in motivation, these works were designed for sum-pooled embedding tables and do not directly translate to the sequence-based DLRMs explored in this work.

\section{Conclusion}
\label{conclusion}

This paper focuses on sharding sequence embeddings for  industry-scale DLRMs. 
We perform an in-depth analysis to model the systems scaling of various sharding strategies used to partition and place these sequence embedding tables across a hardware topology, as well as develop a new sharding strategy which adapts to the heterogeneous network typologies of real-world training clusters. Furthermore, we demonstrate how the systems demand placed upon the hardware actually scales based on the underlying sequence feature distribution. Building upon this, we propose FlexShard, a sharding technique which optimally partitions and places sequence embedding tables at a per-row granularity across multiple potential sharding strategies, for the given hardware topology and memory constraints. We implement and evaluate FlexShard on top of the state-of-the-art open-source DLRM training framework TorchRec using production data. FlexShard is able to achieve an over 85\% reduction in global all-to-all communication traffic, corresponding to an almost 6x speedup in end-to-end communication latency, on representative industry-scale sequence DLRMs.

\bibliography{references}
\bibliographystyle{mlsys2021}

\appendix

\end{document}